# Swarm Intelligence for Multiobjective Optimization of Extraction Process

**T. Ganesan***
*Department of Chemical Engineering,*
*Universiti Technologi Petronas,*
*31750 Tronoh, Perak, Malaysia*
*\*email: tim.ganesan@gmail.com*

**I. Elamvazuthi**
*Department of Electrical & Electronics Engineering,*
*Universiti Technologi Petronas,*
*31750 Tronoh, Perak, Malaysia*

**P. Vasant**
*Department of Fundamental & Applied Sciences,*
*Universiti Technologi Petronas,*
*31750 Tronoh, Perak, Malaysia*

**ABSTRACT**

*Multi objective (MO) optimization is an emerging field which is increasingly being implemented in many industries globally. In this work, the MO optimization of the extraction process of bioactive compounds from the Gardenia Jasminoides Ellis fruit was solved. Three swarm-based algorithms have been applied in conjunction with normal-boundary intersection (NBI) method to solve this MO problem. The gravitational search algorithm (GSA) and the particle swarm optimization (PSO) technique were implemented in this work. In addition, a novel Hopfield-enhanced particle swarm optimization was developed and applied to the extraction problem. By measuring the levels of dominance, the optimality of the approximate Pareto frontiers produced by all the algorithms were gauged and compared. Besides, by measuring the levels of convergence of the frontier, some understanding regarding the structure of the objective space in terms of its relation to the level of frontier dominance is uncovered. Detail comparative studies were conducted on all the algorithms employed and developed in this work.*

Keywords: Multiobjective Optimization; Metaheuristics; Normal Boundary Intersection; Measurement Metrics; Particle Swarm Optimization; Gravitational Search Algorithm.

**INTRODUCTION**

Multi-criteria or multi-objective (MO) scenarios have become increasingly prevalent in industrial engineering environments (Statnikov & Matusov, 1995; Zhang and Li, 2007; Li and Zhou, 2011). MO optimization problems are commonly tackled using the concept of Pareto-optimality to trace-out the non-dominated solution options at the Pareto curve (Zitzler & Thiele, 1998; Deb *et al.*, 2002). Other methods include the weighted techniques which involve objective function aggregation resulting in a master weighted function. This master weighted function is then solved for various weight values (which are usually fractional) (Fishburn, 1967; Triantaphyllou, 2000; Luyben. & Floudas, 1994; Das & Dennis, 1998). Using these techniques, the weights are used to consign relative importance or priority to the objectives in the master aggregate function. Hence, alternative near-optimal solution options are generated for various values of the scalars. In this chapter, the Normal Boundary Intersection (NBI) scheme (Das & Dennis, 1998) was used as a scalarization tool to construct the Pareto frontier. In Sandgren (1994) and Statnikov & Matusov (1995), detail examples and analyses on MO techniques for problems in engineering optimization are presented.

Many optimization techniques have been implemented for solving the extraction process problem (e.g. Hismath *et al.*, 2011; Jie and Wei, 2008). In addition, evolutionary techniques such as DE have also been employed for extraction process optimization (Ubaidullah *et al*., 2012). The MO problem considered in this work was formulated by Shashi *et al*, (2010). This problem involves the optimization of the yields of certain chemical products which are extracted from the Gardenia *Jasminoides Ellis* fruit. The MO optimization model was developed in Shashi *et al*, (2010) to maximize the extraction yields which are the three bioactive compounds; crocin, genipoiside and total phenolic compounds. The optimal extraction parameters which construct the most dominant Pareto frontier are then identified such that the process constraints remain unviolated. In Shashi *et al*., (2010), the MO problem was tackled using the real-coded Genetic Algorithm (GA) approach to obtain a single individual optima and not a Pareto frontier. In that work, measurement metrics were not employed to evaluate the solution quality in detail. In addition, the work done in Shashi *et al*., (2010) focused on modeling the system rather than optimizing it. The authors of that work employed only one optimization technique and did not carry out extensive comparative analysis on the optimization capabilities. Due to this setbacks, these factors are systematically addressed in this chapter to provide some insights on the optimization of the extraction process.

Over the past years, swarm intelligence-based meta-heuristic techniques have been applied with increasing frequency to industrial MO scenarios. Some of the most effective swarm approaches have been devised using ideas from Newtonian gravitational theory (Rashedi *et al*., 2009), dynamics of fish movement (Neshat *et al*., 2012) and birds flocking behaviors (Kennedy & Eberhart, 1995). In this work, three swarm-based techniques; gravitational search algorithm (GSA) (Rashedi *et al*., 2009), particle swarm optimization (PSO) (Kennedy & Eberhart, 1995) and the novel Hopfield-Enhanced PSO (HoPSO) were employed to the extraction problem (Shashi *et al*, 2010). The measurement techniques; the convergence metric (Deb & Jain, 2002) and the Hypervolume Indicator (HVI) (Zitzler & Thiele, 1998) were used to analyze the solution spread produced by these algorithms.

The HVI is a set measure reflecting the volume enclosed by a Pareto front approximation and a reference set (Emmerich *et al*., 2005). The convergence metric on the other hand measures the degree at which the solutions conglomerate towards optimal regions of the objective space. Using the values obtained by the measurement metrics, the correlation between the convergence and the degree of dominance (measured by the HVI) of the solution sets is obtained and discussed. The solutions constructing the Pareto frontier obtained using the developed HoPSO algorithm is also subjected to the analyses mentioned above. In this work, all computational procedures were developed using the Visual C++ Programming Language on a PC with an Intel i5-3470 (3.2 GHz) Processor.

This chapter is organized as follows: Section 2 presents an overview on industrial MO optimization while Section 3 discusses some fundamentals on pareto dominance. Section 4 gives the problem description followed by the NBI approach in Section 5. Section 6 provides some details regarding swarm intelligence techniques employed in this work. Section 7 provides information related to the measurement indicators while in Section 8 the computational results are discussed. Finally, this chapter ends with some conclusions and recommendations for future work.

## INDUSTRIAL MULTIOBJECTIVE OPTIMIZATION

Over the past years, MO optimization has been introduced and applied into many engineering problems. Some of these developments shall be briefly discussed in the following. In Aguirre *et al*., (2004), a MO evolutionary algorithm with an enhanced constraint handling mechanism was used to optimize the circuit design of a Field Programmable Transistor Array (FPTA). The authors used the Inverted Shrinkable Pareto Archived Evolution Strategy (ISPAES) for the MO optimization of the circuit design. Another MO problem involving engineering design was solved by Reddy and Kumar, (2007). In that work, a MO swarm intelligence algorithm was developed by incorporating the Pareto dominance relation into the standard particle swarm optimization (PSO) algorithm. Three engineering design problems, the 'two bar truss design (Palli *et al.*, 1999)', 'I-beam design (Yang *et al.*, 2002)' and the 'welded beam design (Deb *et al.*, 2000)' problems were successfully solved in Reddy and Kumar, (2007). In the area of thermal system design, the MO optimization of an HVAC (Heating, Ventilating, Air-Conditioning and Cooling) system was carried out by Kusiak *et al.*, (2010). In that work, a neural network was used to derive the MO optimization model. This model was then optimized using a multi-objective PSO algorithm (MOPSO). Using this algorithm, the authors identified the optimum control settings for the supply air temperature and static pressure to minimize the air handling unit energy consumption while maintaining air quality. On the other hand, in materials engineering, the MO optimization of the surface grinding process was carried out by Pai *et al.*, (2011). In that work, the enhanced version of the Non-dominated Sorting Genetic Algorithm (NSGA-II) was employed to optimize the machining parameters of the grinding process to increase the utility for machining economics and to increase the product quality (Azouzi & Guillot, 1998). Another application of the NSGA-II algorithm to engineering system design was done in Nain *et al.*, (2010). In Nain *et al.*, (2010), the authors optimized the structural parameters (area and length of the TEC elements) of the Thermoelectric Cooler (TEC). The Coefficient of Performance (COP) and the Rate of Refrigeration (ROR) was successfully maximized in that work.

Recently, MO optimization methods have also penetrated the power and energy industries. For instance, in Van Sickel *et al.*, (2008), the MO optimization of a fossil fuel power plant was done using multi-objective evolutionary programming (MOEP) and the multi-objective particle swarm optimization (MOPSO) algorithms. The MO techniques in that work were done to develop reference governors for power plant control systems. Another work on the MO optimization of reference governor design for power plants was done by Heo and Lee, (2006). In that work, PSO variant algorithms were used to find the optimal mapping between unit load demands and pressure set point of a fossil fuel power plant. By this approach, the optimal set points of the controllers under a large variety of operation scenarios were achieved. Similarly, in the works of Song and Kusiak (2010), temporal processes in power plants were optimized using MO techniques. In that work, the central theme was to maximize the boiler efficiency while minimizing the limestone consumption. Two approaches; the Data Mining (DM) and evolutionary strategy algorithms were combined to solve the optimization model. In Song and Kusiak, (2010) the MO optimization of temporal dependent processes were successfully completed by identifying the optimum control parameters. One other area at which MO optimization has been applied with considerable success is in the field of economic/environmental dispatch for power systems. For instance, in the works of Gunda and Acharjee, (2011), an MO economic/environmental dispatch problem was solved using the Pareto Frontier Differential Evolution (PFDE) approach. By using this technique, the authors managed to minimize the fuel consumption and emissions with minimal energy loss. This triple-objective problem was successfully solved without the violation of the system's security constraints. Another similar problem was tackled in King *et al.*, (2005). In that work, power generation optimization was done to minimize the total fuel costs as well as the amount of emission.

## PARETO FRONTIER AND DOMINANCE LEVELS

To get a clearer picture of the idea the Pareto-compliance, the concepts of Pareto dominance needs to be defined. Pareto dominance can be categorized into three types which are; strictly dominates ($\succ$), weakly dominates ($\succeq$) and indifferent (~). Let two solution vectors be **a** and **b**. Then if the solution vector **a** dominates the vector **b** in all the objectives then **a** strictly dominates **b** (**a** $\succ$ **b**). If the solution vector **a** dominates the vector **b** in some of the objectives but not all, then **a** weakly dominates **b** (**a** $\succeq$ **b**). Finally, if the solution vector **a** does not dominate the vector **b** and the solution vector **b** does not dominate **a** as well in all the objectives, then **a** is indifferent to **b** (**a** ~ **b**). Strictly Pareto-compliant can be defined as the following: Let there be two solution sets say; $X$ and $Y$ for a specific MO problem. If the hypervolume coverage for $X$ is greater than $Y$, then the solution set $X \succ Y$ or $X \succeq Y$. The hypervolume measures the volume of the dominated section of the objective space and can be applied for multi-dimensional scenarios. Implementation of the hypervolume requires a reference point or a 'nadir point'. The nadir point is a point which is dominated by all the solutions from the approximate Pareto frontier. Relative to this point, the volume of the space of all dominated solutions can be computed. A bi-objective depiction of the hypervolume is given in Figure 1:

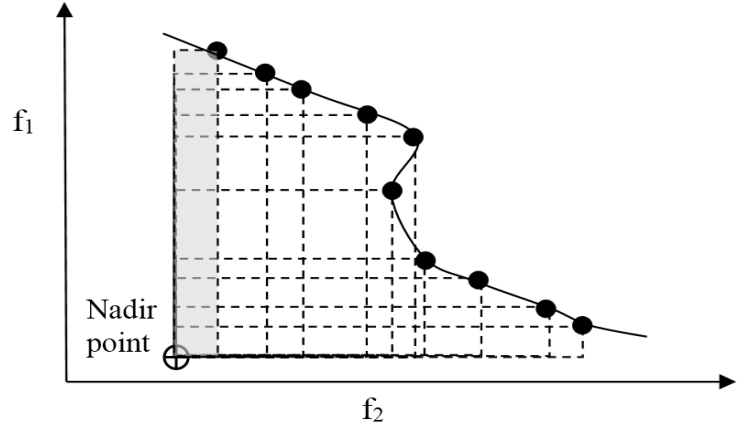

Figure 1: Hypervolume Construction for a Two Objective Maximization Problem.

The larger the value of the hypervolume, the more dominant the solution is in the objective space. The hypervolume is strictly monotonic. Its computational effort is exponential to the amount of solution vectors however requires a bounding vector (nadir point).

## PROBLEM DESCRIPTION

The model for MO problem considered in this work was developed by Shashi *et al.*, (2010). This problem involves the optimization of the yields of certain chemical products which are extracted from the Gardenia *Jasminoides Ellis* fruit. The phenolic compounds in Gardenia *Jasminoides Ellis* have high antioxidant capabilities which make this fruit valuable for medicinal uses (Li *et al.*, 2008). Compared to other natural food pigments, the colouring constituents of the fruit of Gardenia are non-toxic and chemically stable (Van Calsteren *et al.*, 1997) The constituents present in the Gardenia fruit (Osima *et al.*, 1988) are iridoid glycosides (for instance; gardenoside, geniposide, gardoside and scandoside methyl ester). These constituents could be converted into blue colorants under aerobic condition by enzymes or some microorganism. The Gardenia fruit extract in its rudimentary form also contains phenolic compounds high antioxidant capacity in abundance (Li *et al.*, 2008). The MO optimization model in Shashi *et al.*, (2010) was for the extraction process of bioactive compounds from gardenia with respect to the constraints. The MO optimization model was developed to maximize the yield consists of three bioactive compound; crocin ($f_1$), geniposide ($f_2$) and total phenolic compounds ($f_3$). This model is presented as the following:

*Maximize* → *Crocin, $f_1$*
*Maximize* → *Geniposide, $f_2$*
*Maximize* → *Total phenolic compounds, $f_3$*

subject to *process constraints*. (1)

The objective functions represent the yields of each of the bioactive compound in the units of mg/g of dry powder as in Shashi *et al.*, (2010). The objective function modeled with respect to the decision variables are as the following:

$$f_1 = 3.8384907903 + 0.0679672610X_1 + 0.0217802311X_2 + 0.0376755412X_3 \\ - 0.0012103181X_1^2 + 0.0000953785X_2^2 - 0.0002819634X_3^2 + 0.0005496524X_1X_2 \\ - 0.0009032316X_2X_3 + 0.0008033811X_1X_3 \quad (2)$$

$$f_2 = 46.6564201287 + 0.6726057655X_1 + 0.4208752507X_2 + 0.9999909858X_3 \\ - 0.0161053654X_1^2 - 0.0034210643X_2^2 - 0.0116458859X_3^2 + 0.0122000907X_1X_2 \\ - 0.0095644212X_2X_3 + 0.0089464814X_1X_3 \quad (3)$$

$$f_3 = -6.3629169281 + 0.4060552042X_1 + 0.3277005337X_2 + 0.3411029105X_3 \\ - 0.0053585731X_1^2 - 0.0020487593X_2^2 - 0.0042291040X_3^2 + 0.0017226318X_1X_2 \\ - 0.0011990977X_2X_3 + 0.0007814998X_1X_3 \quad (4)$$

The decision variables (which the process extraction parameters) are constrained as per the experimental setup described in Shashi *et al.*, (2010). The constraints are as the following:

$$X_1 \in [19.5, 80.5] \\ X_2 \in [27.1, 72.9] \\ X_3 \in [7.1, 52.9] \quad (5)$$

where $X_1$ is the concentration of ethanol in %, $X_2$ is the extraction temperature in $^oC$ and $X_3$ is the extraction time in minutes.

## NORMAL BOUNDARY INTERSECTION METHOD

The NBI method (Das and Dennis, 1998) is a geometrically-inspired scalarization technique for solving MO optimization problems. As opposed to the Weighted Sum Method, the NBI method searches for a near-uniform spread of Pareto-optimal solutions in the frontier. This makes the NBI approach a more interesting alternative as compared to the Weighted Sum Method when solving non-convex MO problem. Given a generic triple-objective MO optimization problem:

$$Min_{x \in X} \ F(x) \quad s.t.$$

$$X = \{x : g(x) = 0; h(x) \leq 0, x \in [1,3]\}$$

$$F^* = \begin{pmatrix} f_1^* & f_2^* \end{pmatrix}^T \quad (6)$$

where $F^*$ is the utopia point for this MO problem. The individual minimum denoted as $x^*_i$ is obtained for $i \in [1,3]$. This way the the simplex for the convex hull of the individual minima is generated:

$$\Gamma = \{\varphi \cdot Y : \varphi = F(x_i^*); Z = \beta_i : i \in [1,3]\} \quad (7)$$

where $\varphi$ forms a 3 x 3 matrix and $\sum_{i=1}^{3}\beta_i =1$. Thus, the $\beta$ -sub problem is then constructed as follows:

$$Max_{(X,t)} t \quad s.t.$$
$$\varphi \cdot Z + tn = F(x) \quad and \quad x \in X \tag{8}$$

where *t* is the distance parameter, and *n* is the normal vector at the point towards the utopia point. Thus, the NBI approach aims to search for the maximum distance, *t* in the direction of the normal vector, *n* . This distance, *t* is thus between a point on the simplex and the utopia point. Then the scalar values, *Z* are then varied systematically to develop a near-uniform spread of the Pareto frontier.

## SWARM INTELLIGENCE

In the field of optimization, an artificial swarm is a group of virtual organisms or agents that behave interactively to achieve some pre-defined goal. This form of interaction gives the individuals higher capabilities as well as better efficiency in achieving the goal as compared to a single individual. In this work, two conventional swarm-based techniques (GSA and PSO) are employed. In addition, an enhanced version of the conventional PSO approach (Ho-PSO) is developed and implemented on the extraction process problem.

### 1. Particle Swarm Optimization

The PSO algorithm was originally proposed by Kennedy and Eberhart (1995). This technique springs from two distinct frames of notions. The first notion was based on the investigation of swarming or flocking behaviors among some species of organisms (such as; birds, fishes, etc.). The second idea was inspired by the field of evolutionary computing. The PSO algorithm operates by exploring the search space for optimal candidate solutions and then evaluates these solutions with respect to some user defined fitness condition. The candidate optimal solutions obtained by this algorithm are achieved as a result of particles which are swarming through the fitness landscape. The velocity and position updating rule is critical to the optimization capabilities of this technique. The velocity and the position of each particle are updated using the following equations:

$$v_i(t+1) = wv_i(t) + c_1 r_1 [\hat{x}_i(t) - x_i(t)] + c_2 r_2 [g(t) - x_i(t)] \tag{9}$$
$$x_i(t+1) = x_i(t) + v_i(t+1) \tag{10}$$

where each particle is identified by the index *i*, $v_i(t)$ is the particle velocity and $x_i(t)$ is the particle position with respect to iteration *(t)*. The terms, $c_1 r_1 [\hat{x}_i(t) - x_i(t)]$ and $c_2 r_2 [g(t) - x_i(t)]$ represent the personal and social influence respectively. These terms control the level of exploration and exploitation of the PSO algorithm during the search process in the objective space. The parameters *w*, $c_1$ and $c_2$ are defined by the user while $r_1$ and $r_2$ are random variables. These parameters are typically constrained with the following ranges:

$$w \in [0,1.2], c_1 \in [0,2], c_2 \in [0,2], r_1 \in [0,1], r_2 \in [0,1]. \tag{11}$$

The stopping criterion is satisfied when all particles/candidate solutions have reached their fittest positions during the iterations. The execution procedures of the PSO technique and the initial parameters used in this work are shown in Algorithm 1 and Table 1 respectively.

**Algorithm 1**: Particle Swarm Optimization (PSO) Algorithm
**Step 1**: Initialize no of particles, $i$ and the algorithm parameters $w, c_1, c_2, r_1, r_2, n_o$
**Step 2**: Randomly set initial position $x_i(t)$ and velocity, $v_i(t)$
**Step 3**: Compute individual and social influence
**Step 4**: Compute position $x_i(t+1)$ and velocity $v_i(t+1)$ at next iteration
**Step 5**: If the swarm evolution time, $t > t_o + T_{max}$, update position $x_i$ and velocity $v_i$ and go to Step 3, else proceed to Step 6
**Step 6**: Evaluate fitness swarm.
**Step 7**: If fitness criterion satisfied, halt and print solutions, else go to step 3.

*Table 1: PSO Parameter Settings*

| Parameters | Values |
| --- | --- |
| Initial parameter ($c_1, c_2, w$) | (1, 1.2, 0.8) |
| Number of particles | 6 |
| initial social influence ($s_1, s_2, s_3, s_4, s_5, s_6$) | (1.1, 1.05, 1.033, 1.025, 1.02, 1.017) |
| initial personal influence ($p_1, p_2, p_3, p_4, p_5, p_6$) | (3, 4, 5, 6, 7, 8) |

In the event the position of all the particles converges during the iterative process, the solutions are considered feasible if:

- The specified ranges are respected and no constraint violation exists.
- All the decision variables are non-negative
- No further optimization of the objective function occurs

Therefore, based on these criteria it can be said that the fitness requirements have been met. The candidate solutions are hence at their fittest and in effect the program is stopped and the solutions are printed.

## 2. Gravitational Search Algorithm

The GSA algorithm is a meta-heuristic algorithm first developed by Rashedi *et al.*, (2009). This technique was developed by the using the gravitational laws and the idea of masses interactions as the basis. This algorithm uses the Newtonian gravitational laws where the search agents act as masses. Therefore, the gravitational forces influence the motion of these masses, where lighter masses gravitate towards the heavier masses during these interactions which signify the algorithm's progression towards the optima. The GSA algorithm randomly generates a distribution of masses, $m_i(t)$ and also sets the position for these masses, $x_i^d$ at the initial stages. For a minimization problem, the least fit mass, $m_i^{worst}(t)$ and the fittest mass, $m_i^{best}(t)$ at time $t$ with $N$ number of masses are computed as follows:

$$m^{best}(t) = \min_{j \in [1,N]} m_j(t) \tag{12}$$

$$m^{worst}(t) = \max_{j \in [1,N]} m_j(t) \tag{13}$$

For a maximization problem, the operators are swapped for the best and worst masses respectively. The inertial mass, $m_i'(t)$ and gravitational masses, $M_i(t)$ are then computed based on the fitness map developed previously.

$$m'_i(t) = \frac{m_i(t) - m^{worst}(t)}{m^{best}(t) - m^{worst}(t)} \tag{14}$$

$$M_i(t) = \frac{m_i(t)}{\sum_{j=1}^{N} m_j(t)} \tag{15}$$

such that,

$$M_{ai} = M_{pi} = M_{ii} = M_i : i \in [1, N] \tag{16}$$

Then the gravitational constant, $G(t+1)$ and the Euclidean distance $R_{ij}(t)$ is computed as the following:

$$G(t+1) = G(t) \exp\left(\frac{-\alpha t}{T_{max}}\right) \tag{17}$$

$$R_{ij}(t) = \sqrt{(\Delta x_i(t))^2 - (\Delta x_j(t))^2} \tag{18}$$

where $\alpha$ is some arbitrary constant and $T_{max}$ is the maximum number of iterations, $x_i(t)$ and $x_j(t)$ are the positions of particle $i$ and $j$ at time $t$. The interaction forces at time $t$, $F_{ij}^d(t)$ for each of the masses are then computed:

$$F_{ij}^d(t) = G(t)\left(\frac{M_{pi}(t) \times M_{aj}(t)}{R_{ij}(t) + \varepsilon}\right) \times \left(x_j^d(t) - x_i^d(t)\right) \tag{19}$$

where $\varepsilon$ is some small parameter. The total force acting on each mass $i$ is given in a stochastic form as the following:

$$F_i^d(t) = \sum_{\substack{j=1 \\ i \neq j}}^{N} rand(w_j) F_{ij}^d(t) : rand(w_j) \in [0,1] \tag{20}$$

where $rand(w_j)$ is a randomly assigned weight. Consequently, the acceleration of each of the masses, $a_i^d(t)$ is then as follows:

$$a_i^d(t) = \left(\frac{F_i^d(t)}{M_{ii}(t)}\right) \tag{21}$$

After the computation of the particle aceleration, the particle position and velocity is then calculated:

$$v_i^d(t+1) = rand(w_j) + v_i^d(t) + a_i^d(t) \tag{22}$$

$$x_i^d(t+1) = x_i^d(t(t)) + v_i^d(t(t+1)) \tag{23}$$

where *rand(w_j)* is a randomly assigned weight. The iterations are then continued until the all mass agents are at their fittest positions in the fitness landscape and some stopping criterion which is set by the user is met. The GSA algorithm is presented in Algorithm 2 and the parameter settings are given in Table 3:

**Algorithm 2**: Gravitational Search Algorithm (GSA)
**Step 1**: Initialize no of particles, $m_i$ and initial positions, $x_i(0)$
**Step 2**: Initialize parameters: $G(0)$, $\alpha$.
**Step 3**: Compute gravitational & inertial masses based on the fitness map
**Step 4**: Compute the gravitational constant, $G(t)$
**Step 5**: Compute distance between agents, $R_{ij}(t)$
**Step 6**: Compute total force, $F_i^d(t)$ and the acceleration $a_i^d(t)$ of each agent.
**Step 7**: Compute new velocity $v_i(t)$ and position $x_i(t)$ for each agent
**Step 8**: If the fitness criterion is satisfied and $t= T_{max}$, halt and print solutions
    else proceed to step 3

*Table 2: GSA Parameter Settings*

| Parameters | Values |
|---|---|
| Initial parameter ($G_o$) | 100 |
| Number of mass agents, $n$ | 6 |
| Constant parameter, $\alpha$ | 20 |
| Constant parameter, $\varepsilon$ | 0.01 |

## 3. Hopfield-Enhanced Particle Swarm Optimization

The HoPSO employed in this work merges the ideas from the Ising spin models (Amit *et al.*, 1986), the Hopfield Neural Networks (HNN) (Hopfield, 1982; Hopfield, 1984) and the PSO technique. The Ising model is a ferromagnetism model in statistical physics developed by Ernst Ising in 1925 (Dyson, (1969)). This model is constructed based on the concept that atomic configurations can be represented in terms of magnetic dipole moments (atomic spins – the quantization of magnetism) which are in either in state +1 or -1. These spin interactions are localized to their closest neighbour and are usually in a lattice arrangement. The central idea of the Ising model is to detect phase transitions in real substances. Hence, repetitive magnetization would under certain circumstances and after a period of time would cause the total energy of the magnetized material to converge into local minima.

The statistical physics models developed in the Ising model (Dyson, 1969) then inspired the development of a new type of neural net called the Hopfield Recurrent Artificial Neural Network (HNN) with improved convergence properties. The HNN was developed in 1982 by Hopfield, (1982) and Hopfield, (1984). These neural nets observed to have applications in optimization problems (for instance in Lee, Sode-Yome *et al.*, (1998) and Tank & Hopfield (1986)).

One of the key features of the HNN is that there is a decrease in the energy by a finite amount whenever there is a change in the network's state. This essential property confirms or accentuates the convergence of the output whenever the network state is changed. HNNs are usually constructed by a finite number of interlinked neurons. These neurons update their weights or their activation values (outputs from threshold neurons) independently relative to other neurons in the network. It is important to take note that the neurons in these sorts of networks are not directly connected to themselves and each neurons function as an input as well as output. In HNNs, the activation values are usually binary (+1 or -1) and all the weights of neurons are symmetric ($w_{ij} = w_{ji}$).

In this work, the ideas of discreet magnetic spin as well symmetric weight assumptions (from the Ising spin model) that were used in the HNN are applied to the PSO algorithm to improve its convergence capabilities. Similarly, a set of random weights $w_{ij}$ was initialized and the symmetric property was imposed. Then, the modification of the particle position update equation in the PSO algorithm was done:

$$x_i(t+1) = x_i(t) + v_i(t) \tag{24}$$

where $x_i(t)$ is the particle position and $v_i(t)$ is the particle velocity and $t$ is the iteration counter. The weight previously defined is then introduced as a coefficient that dampens the position of the previous iteration such as the following:

$$s_j(t+1) = w_{ij}(t)x_i(t) + v_i(t) \tag{25}$$

The particle position at the next iteration is computed in the following piece-wise form:

$$x_j(t+1) = \begin{cases} +1 & \text{iff} \quad s_j(t+1) > U \\ -1 & \text{iff} \quad s_j(t+1) < U \\ x_j(t) & \text{iff} \quad \text{otherwise} \end{cases} \tag{26}$$

where $U$ is the threshold which user-defined. Then the energy function is defined as follows:

$$E = -\frac{1}{2}\sum_{j=1}\sum_{i=1} x_j(t+1)x_j(t)w_{ij} - \sum_{i=1} \theta x_j(t) \quad \text{such that} \quad j \neq i \tag{27}$$

where $\theta$ is the coefficient which user-defined. This way the PSO algorithm is iterated until the total energy of the system reaches its local minima. To detect the instance when the HoPSO has reached the local minima, the difference between the energy levels at two consequent states is computed as follows:

$$dE = E_{n+1} - E_n \tag{28}$$

where $n$ is the index that denotes the states. Thus, a new variant for the PSO algorithm is developed which in this work shall be termed as Hopfield PSO (HoPSO) algorithm. The algorithm and the flowchart for this HoPSO method are given in Algorithm 3 and Figure 2 respectively:

**Algorithm 3:** Hopfield Particle Swarm Optimization (HoPSO)

**Step 1**: Set no of particles, $i$ and the initialize parameter settings $w, c_1, c_2, r_1, r_2, n_o$
**Step 2**: Randomly initialize particles' position $x_i(t)$ and velocity $v_i(t)$
**Step 3**: Randomly initialize weights, $w_{ij}(t)$
**Step 4**: Enforce symmetry condition on weights
**Step 5**: Calculate cognitive and social components of the particles
**Step 6**: Compute position $x_i(t+1)$ and velocity $v_i(t+1)$ of the particles at next iteration
**Step 7**: Compute energy function
**Step 8**: Proceed with the evaluation of the fitness of each particle in the swarm.
**Step 8**: If the energy difference between states are < 0, proceed to Step 9
      else go to Step3.
**Step 9**: If the fitness conditions are satisfied and $t < T_{max}$, stop program and print solutions,
      else go to Step 3.

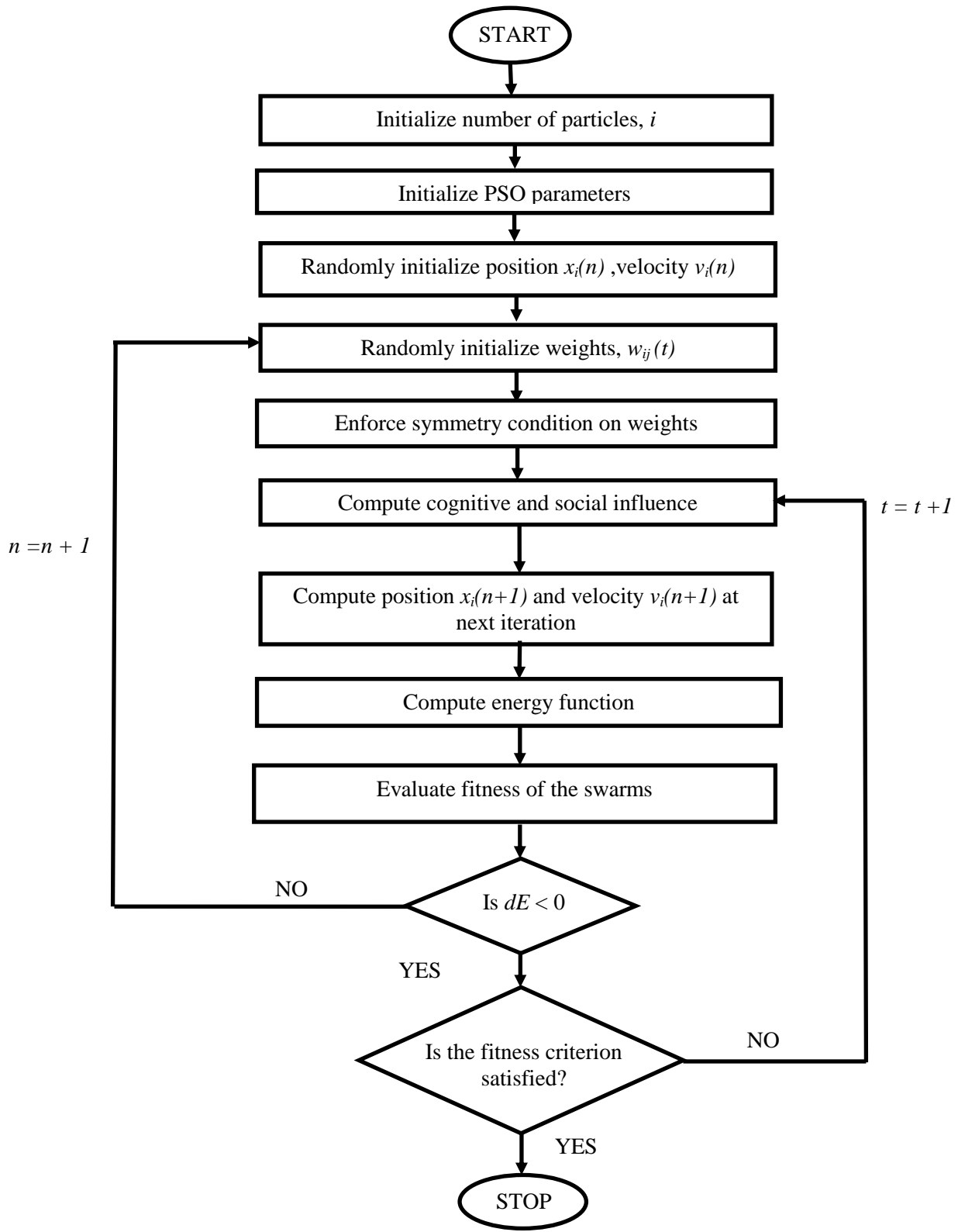

Figure 2: Flowchart for the HoPSO technique.

The parameter settings specified in the HoPSO algorithm in this work is as in Table 4:

*Table 3: HoPSO Parameter Settings*

| Parameters | Values |
| --- | --- |
| Initial parameter ($c_1$, $c_2$, $r_1$, $r_2$, $w$) | (1, 1.2, 0.5, 0.5, 0.8) |
| Number of particles | 6 |
| Initial social influence ($s_1$, $s_2$, $s_3$, $s_4$, $s_5$, $s_6$) | (1.1, 1.05, 1.033, 1.025, 1.02, 1.017) |
| Initial personal influence ($p_1$, $p_2$, $p_3$, $p_4$, $p_5$, $p_6$) | (3, 4, 5, 6, 7, 8) |
| Constant Parameter, $U$ | 100 |
| Constant Parameter, $\theta$ | 0.02 |

## MEASUREMENT INDICATORS

### 1. Convergence Metric

The convergence metric (Deb & Jain, 2002) was developed to measure the degree of convergence of a solution set relative to a reference/basis set. Since the Pareto optimal frontier was not known, a target vector $P^*$ which is the most dominant vector was employed as the reference set. For a set of solutions, the formulation to compute the convergence metric for a single run of the program is as follows:

$$d_i = \min_{j=1}^{|P^*|} \sqrt{\sum_{k=1}^{M}\left(\frac{f_k^i - f_k^j}{f_k^{\max} - f_k^{\min}}\right)} \tag{29}$$

where $i$ and $j$ denotes the subsequent objective function values, $k$ is the index which denotes the objective function, $f_k^{max}$ is the maximum objective function value, $f_k^{min}$ is the minimum objective function value and $M$ denotes the overall number of objectives. For this convergence metric, low the metric values indicate high convergence characteristics among the solution vectors.

### 2. Hypervolume Indicator

One approach that has been effective in measuring the quality of the solution set that constructs the Pareto-frontier in cases where the Pareto frontier is unknown is the Hypervolume Indicator (HVI) (Zitzler & Thiele, 1998). The HVI is a Pareto-compliant gauge that is employed for the measurement of the quality of solution sets in MO optimization problems (Zitzler & Thiele, 1998; Beume *et al.*, 2007). Pareto-compliance is defined as follows:

*Let two solution sets exists such that these sets are produced by some arbitrary algorithm for a particular MO problem. Then the property of Pareto compliance implies that the indicator would consistently produce a higher value for one of the solution set only if that solution set strictly dominates the other.*

In the past couple of years, many investigations were conducted on methods to apply evolutionary algorithms to MO problems. Besides, many ideas have also been proposed on methods to attain a good solution spread over the Pareto frontier. Unfortunately, in all these works, the optimization goal remains fuzzy since there is no established technique to measure the quality of a solution set produced by any MO algorithm (especially in cases where there are more than two objectives). The impact of this issue magnifies especially in real-world problems where it is often that the Pareto frontier is unknown (unlike theoretical test functions where the Pareto frontier is known and can be used to benchmark the solution quality).

Recently, this indicator has been frequently applied in many works involving MO problems. The HVI is the only indicator which is strictly Pareto-compliant that can be used to measure the quality of solution sets in MO optimization problems. To get a clearer picture of the idea the Pareto-compliance, the concepts of Pareto dominance needs to be defined.

Pareto dominance can be categorised into three types which are; strictly dominates ($\succ$), weakly dominates ($\succeq$) and indifferent (~). Let two solution vectors be **a** and **b**. Then if the solution vector **a** dominates the vector **b** in all the objectives then **a** strictly dominates **b** (**a** $\succ$ **b**). If the solution vector **a** dominates the vector **b** in some of the objectives but not all, then **a** weakly dominates **b** (**a** $\succeq$ **b**). Finally, if the solution vector **a** does not dominate the vector **b** and the solution vector **b** does not dominate **a** as well in all the objectives, then **a** is indifferent to **b** (**a** ~ **b**).

Strictly Pareto-compliant can be defined as the following. Let there be two solution sets say; $X$ and $Y$ for a particular MO problem. If the HVI value for $X$ is greater than $Y$, then the solution set $X \succ Y$ or $X \succeq Y$. The HVI measures the volume of the dominated section of the objective space and can be applied for multi-dimensional scenarios. When using the HVI, a reference point needs to be defined. In this work, we define a 'nadir point'. The nadir point is a point which is dominated by all the solutions from the approximate Pareto frontier. Relative to this point, the volume of the space of all dominated solutions can be computed. The HVI of a solution set $x_d \in X$ can be defined as follows:

$$HVI(X) = vol\left( \bigcup_{(x_1,...x_d) \in X} [r_1, x_1] \times ... \times [r_d, x_d] \right) \tag{30}$$

where $r_1,...,r_d$ is the reference point and $vol(.)$ being the usual Lebesgue measure. The larger the value of the HVI, the more dominant the solution is in the objective space. The characteristics of the HVI are as follows:

a) Strictly monotonic indicator.
b) Its computational effort is exponential to the amount of solution vectors.
c) It requires an upper-bounding vector (nadir point).

In this work the HVI is used to measure the quality of the approximation of the Pareto front by the solution sets produced by the algorithms when used in conjunction with the NBI approach. In this work the HVI is used to measure the quality of the approximation of the Pareto front by the swarm algorithms when used in conjunction with the NBI framework.

## COMPUTATIONAL RESULTS & ANALYSIS

In this work, the Pareto frontier is constructed using the solution set produced by the swarm algorithms. In this work, 27 solution points were produced by all the algorithms for the construction of the Pareto frontier. The algorithms presented in this work were executed on a 2GHz Intel dual-core processor. The dominance levels of these solutions were measured using the HVI. The nadir point selected in this work is $(r_1, r_2, r_3) = (0, 0, 0)$. The individual solutions with their designated weights produced by the GSA and the PSO algorithms were gauged with the HVI and the ranked solutions were determined. These individual solutions for both the algorithms and their respective dominance levels are shown in Table 4 and 5 respectively.

*Table 4. Ranked Individual Solutions Generated by the GSA Algorithm*

| Description | | Best | Median | Worst |
|---|---|---|---|---|
| Objective Function | $f_1$ | 6.13334 | 5.90007 | 5.88888 |
| | $f_2$ | 81.1329 | 75.1058 | 74.8027 |
| | $f_3$ | 11.9987 | 9.8936 | 9.78725 |
| Decision Variable | $x_1$ | 19.5 | 19.5 | 19.5 |
| | $x_2$ | 27.1 | 27.1 | 27.1 |
| | $x_3$ | 17.0987 | 7.5167 | 7.11013 |
| HVI | | 5970.741 | 4384.15 | 4311.324 |

*Table 5. Ranked Individual Solutions Generated by the PSO Algorithm*

| Description | | Best | Median | Worst |
|---|---|---|---|---|
| Objective Function | $f_1$ | 8.53054 | 8.5228 | 8.52069 |
| | $f_2$ | 108.941 | 108.861 | 108.842 |
| | $f_3$ | 24.7943 | 24.7803 | 24.7694 |
| Decision Variable | $x_1$ | 49.6226 | 49.4068 | 49.4391 |
| | $x_2$ | 72.8999 | 72.7734 | 72.6776 |
| | $x_3$ | 30.6499 | 30.6636 | 30.5293 |
| HVI | | 23041.98 | 22991.18 | 22971.36 |

The associated weights for the best, median and worst solution for the GSA and PSO algorithms are shown in Table 6:

*Table 6. Ranked Individual Solutions Generated by the GSA and PSO Algorithms*

| Weights | | Best | Median | Worst |
|---|---|---|---|---|
| GSA | $w_1$ | 0.4 | 0.4 | 0.1 |
| | $w_2$ | 0.2 | 0.3 | 0.4 |
| | $w_3$ | 0.4 | 0.3 | 0.5 |
| PSO | $w_1$ | 0.5 | 0.3 | 0.2 |
| | $w_2$ | 0.2 | 0.4 | 0.6 |
| | $w_3$ | 0.3 | 0.3 | 0.2 |

The depiction of the solution spread of the approximate Pareto frontier obtained using the GSA and PSO algorithms are shown in Figures 3 and 4:

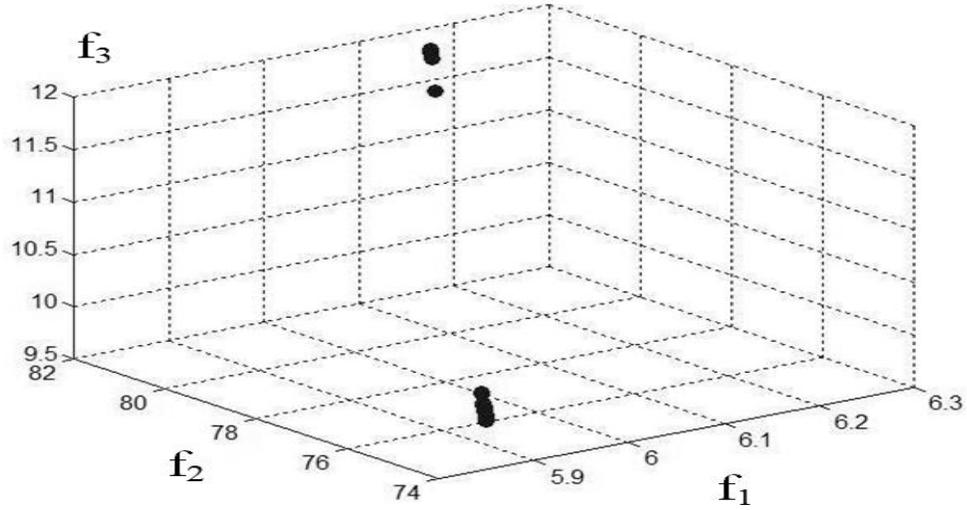

Figure 3: Pareto frontier of the objectives obtained by the GSA method

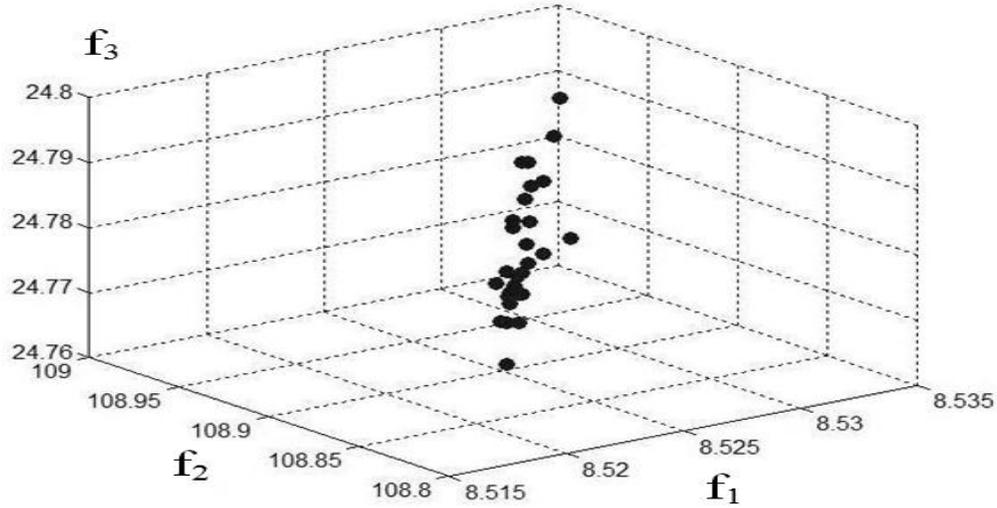

Figure 4: Pareto frontier of the objectives obtained by the PSO technique

Similar to the GSA and PSO algorithms, the individual solutions for the HoPSO technique and the respective dominance levels are given in Table 7.

*Table 7. Ranked Individual Solutions Generated by the HoPSO Algorithm*

| Description | | Best | Median | Worst |
|---|---|---|---|---|
| Objective Function | $f_1$ | 8.5675 | 8.54683 | 8.56125 |
| | $f_2$ | 109.929 | 109.903 | 108.326 |
| | $f_3$ | 24.8 | 24.6649 | 24.1797 |
| Decision Variable | $x_1$ | 55.6416 | 57.699 | 55.3454 |
| | $x_2$ | 72.8689 | 72.5172 | 72.81 |
| | $x_3$ | 33.5713 | 36.9105 | 22.7159 |
| HVI | | 23357.05 | 23168.29 | 22424.4 |

The associated weights ($w_1$, $w_2$, $w_3$) for the best, median and worst solutions produced by the HoPSO technique are (0.5, 0.4, 0.1), (0.1, 0.2, 0.7) and (0.4, 0.4, 0.2). The approximate Pareto frontiers obtained using the HoPSO algorithm is given in Figure 5:

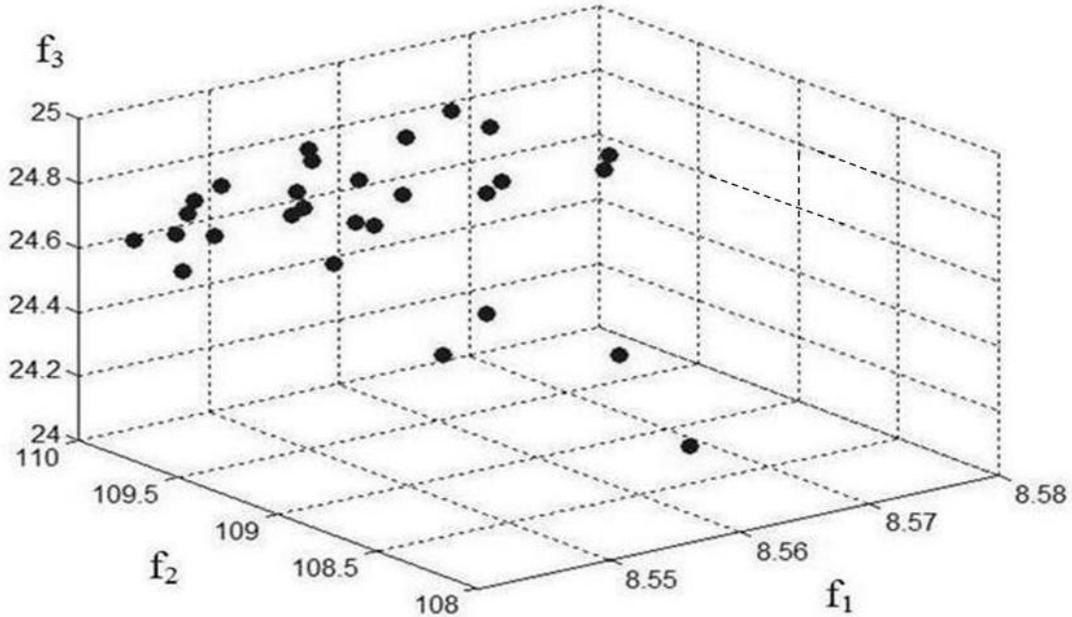

Figure 5: Pareto frontier of the objectives obtained by the HoPSO technique

For comparing the percentage of dominance among the techniques, a simple error metric (%) is utilized. The following equation provides the difference (%) of the performance of Algorithm A when compared against that of Algorithm B using the HVI:

$$\% \ of \ Difference = \left| \frac{HVI_A - HVI_B}{HVI_B} \right| \times 100\% \tag{28}$$

It can be observed using the HVI that the best solution obtained by HoPSO algorithm dominates the best solution produced by the PSO and GSA algorithms by 1.367% and 291.192%. The comparison of the best candidate solutions obtained by the methods employed in this work against the PSO method in Shashi *et al.*, (2010) is shown in Table 8.

*Table 8. The Comparison of the Best Solutions Obtained by the Algorithms*

| Description | | Real-coded GA (Shashi et al., 2010) | PSO | GSA | HoPSO |
|---|---|---|---|---|---|
| Objective Function | $f_1$ | 8.43 | 8.53054 | 6.13334 | 8.5675 |
| | $f_2$ | 110.026 | 108.941 | 81.1329 | 109.929 |
| | $f_3$ | 24.81 | 24.7943 | 11.9987 | 24.8 |
| Decision Variable | $x_1$ | 56.62 | 49.6226 | 19.5 | 55.6416 |
| | $x_2$ | 72.9 | 72.8999 | 27.1 | 72.8689 |
| | $x_3$ | 34.76 | 30.6499 | 17.0987 | 33.5713 |
| HVI | | 23011.75 | 23041.98 | 5970.741 | 23357.05 |

In Table 8, it can be observed that the best solutions produces by HoPSO and PSO algorithms used in this work are more dominant than the real-coded GA approach (Shashi *et al.*, 2010) by 1.5005% and 0.1314% respectively . The HVI computed for the entire frontier of each solution set produced by an algorithm gives the true measure of dominance when compared with another algorithm. In this work, the HVI for the entire frontier was computed for each of the algorithm. The execution time for the GSA, PSO and the HoPSO algorithms to generate the entire frontier are 9.588, 13.113 and 13.268 seconds respectively. The HVI for the entire frontier for the solution sets produced by all the algorithms in this work is shown in Figure 6:

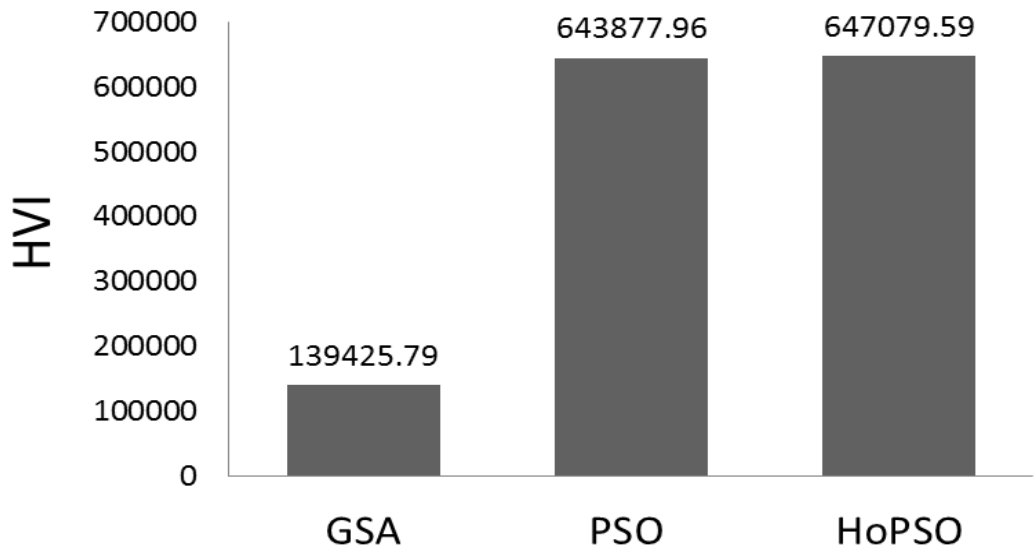

Figure 6: Dominance comparison between Pareto frontiers produced by the GSA, PSO and HoPSO techniques

The Pareto frontier produced by the HoPSO algorithm is more dominant than the one produced by the GSA and PSO algorithms by 364.1% and 0.497% respectively. The level of convergence of the Pareto frontier for all the algorithms employed in this work is measured using the convergence metric (refer to equation 26). The level of convergence of the Pareto frontiers generated by the GSA, PSO and the HoPSO technique is shown in Figure 7:

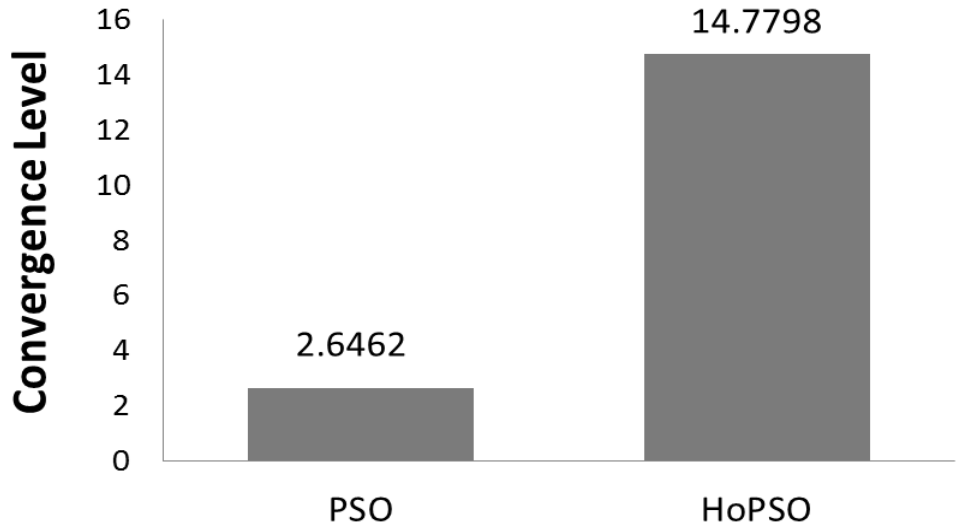

Figure 7: Comparison of the convergence level between Pareto frontiers produced by the PSO and HoPSO techniques

As seen in Table 7, the individual best solution of the HoPSO technique maximizes all the objectives $f_1$, $f_2$ and $f_3$ very effectively as compared to the solutions obtained by using the GSA and PSO algorithms. Thus, it can be said that the HoPSO method in this work outweighs the overall optimization capabilities of PSO, GSA and the real-coded GA (Shashi *et al*., 2010). The HVI values are show in Table 7. The HoPSO technique takes the longest computational time to produce the entire Pareto frontier followed by the PSO and GSA methods. The HoPSO algorithm produces Pareto frontier with the highest level of dominance. However, due to high algorithmic complexity it sacrifices computational time the most as compared to the GSA and PSO methods. As can be seen in Figures 4 and 5, the frontiers produced by the PSO and HoPSO algorithms are more uniform and diversely spaced in the objective space as compared to the GSA algorithm in Figure 3. The frontier produced by the GSA algorithm seems to be confined to certain regions in the objective space. Thus, the solution spacing seems to heavily influence the ability of the algorithm to approximate the Pareto frontier dominantly. Localized solutions on the frontier such as the ones produced by the GSA algorithm omit some of the solutions in the objective space. This in effect causes the GSA algorithm to have a lower level of dominance as compared to the PSO and HoPSO algorithms.

In Figure 7, it can be observed that the Pareto frontier produced by the HoPSO method has a higher level of convergence as compared to the PSO technique. In this case, it could be inferred that the level of convergence as well as the level of dominance of the HoPSO algorithm is higher than that of the PSO technique. Therefore, in a post-analytical sense, it can be said that the level of convergence may be one of the properties of the objective space which is interlinked to the level of dominance.

## CONCLUSIONS

In this work, a new optimal solution as well as a dominant Pareto frontier was achieved by the HoPSO method (refer to Table 7 and Figure 6). The HoPSO technique outperforms the GSA, PSO and the real-coded GA (Shashi *et al*, 2010) methods. Thus, it can be said that the HoPSO method in this work outweighs the overall optimization capabilities of GSA and PSO techniques (refer to levels of dominance in Figure 6). In this work, it can be seen that the level of convergence is interconnected to the level of Pareto dominance. Due to this, the Hopfield enhancement in the HoPSO technique improves the generation of convergent solution which in effect increases the dominance level of the solutions. In this work, the swarm intelligence algorithms performed stable computations during the program executions.

All Pareto-efficient solutions produced by the algorithms developed in this work were feasible and no constraints were compromised. One of the advantages of using the swarm-based algorithms as compared to the other algorithms used in this work is that it produces highly effective results in terms of approximating the Pareto frontier and is computationally efficient.

## FUTURE RESEARCH DIRECTIONS

In future works, other measurement metrics such as spacing or diversity metrics could be employed and the relations between the levels of dominance (using the HVI (Ganesan *et al*., 2013) could be obtained. As mentioned in Ganesan *et al*., (2013), understanding the structure of the objective space is critical when developing algorithmic enhancements.  Keeping this purpose in mind, testing MO with a wide range of algorithms and measurement metrics would improve our understanding on the problem morphology which would in effect provide insight on approaches to enhance our algorithms to achieve the global or near-global optima.

In addition, hybridization procedures (Elamvazuthi *et al*., 2011) could also be utilized and compared with algorithmic enhancement approaches. This way, critical differences in terms of performance and solution quality may be identified. Besides, the approach employed in this work could also be extended towards problems containing various uncertainties (Vasant, *et al*., 2010; Ganesan *et al*., 2014).

## ACKNOWLEDGEMENTS

The authors are very grateful to the reviewers for their efforts in providing useful and constructive insights that has been overlooked by the author during the preparation of the manuscript. The authors would also like to thank Universiti Teknologi Petronas for their support during progress of this work. Special thanks to our friends, colleagues and students who continuously inspire and support us throughout the phases of our research.

## ADDITIONAL READING

# LIST OF SYMBOLS

| Symbols | Description |
|---:|:---|
| $G(t)$ | Gravitational constant (GSA) |
| $n$ | Number of mass agents (GSA) |
| $\alpha$, $\varepsilon$ | Constant parameter (GSA) |
| $R_{ij}(t)$ | Distance between agents (GSA) |
| $F_i^d(t)$ | Total force of each agent (GSA) |
| $a_i^d(t)$ | Acceleration of each agent (GSA) |
| $v_i(t)$ | Velocity of each mass agent (GSA) |
| $x_i(t)$ | Position of each mass agent (GSA) |
| $T$ | Function Evaluations (GSA) |
| $w, c_1, c_2, r_1, r_2$ | PSO Parameters (PSO) |
| $v_i(t)$ | Velocity of each particle (PSO) |
| $x_i(t)$ | Position of each particle (PSO) |
| $T_{max}$ | Maximum limit of function evaluations |
| $s_i$ | Initial Social Influence (PSO) |
| $p_i$ | Initial Individual Influence (PSO) |
| $\phi = F^*$ | Utopia point (NBI) |
| $M$ | Maximum number of objectives (NBI) |
| $X^*_i$ | Individual Minimum (NBI) |
| $g(x)$ | Equality Constraints (NBI) |
| $h(x)$ | Inequality Constraints (NBI) |
| $\beta \in Y$ | Scalar constants (NBI) |
| $t'$ | Distance Parameter (NBI) |
| $\vec{n}$ | Normal Vector (NBI) |
| $w_{ij}$ | Weights of neurons |
| $y_j(t)$ | Activation values from the output |
| $\theta$ | Relaxation Constant |
| $U$ | Threshold value |
| $E$ | Energy of the system |
| $dE$ | Energy Differential |
| $A_i(t)$ | Transfer Function |
| $s_i(t)$ | Piece-wise threshold function |
| $P^j$ | Pareto frontier |
| $c^j, d^j$ | Real-valued scalar |
| $vol(.)$ | Lebesgue measure |
| $HVI(.)$ | Hypervolume indicator |

## KEY TERMS & DEFINITIONS

- Multiobjective optimization

Optimization problems which are represented with more than one objective functions.

- Metaheuristics

A framework consisting of a class of algorithms employed to find good solutions to optimization problems by iterative improvement of solution quality.

- Normal Boundary Intersection

A solution method where a multiobjective optimization problem is geometrically reduced to a beta-sub problem and then solved as a weighted single objective optimization problem.

- Measurement metrics

Mathematical metrics used to identify and measure solution characteristics such as; degree of convergence, diversity and dominance.

- Particle Swarm Optimization (PSO)

A type of metaheuristic algorithm that uses concepts from swarming behavior of organisms to search for optimal solutions.

- Gravitational Search Algorithms (GSA)

A type of metaheuristic algorithm integrates ideas from Newtonian gravitational laws to search for optimal solutions in the objective space.

- Hopfield-Enhanced Particle Swarm Optimization (HoPSO)

A novel swarm intelligence technique which integrates concepts from the Hopfield Neural Nets and swarming behavior of organisms to improve the solution dominance of the standard PSO when solving multiobjective problems.

- Gardenia *Jasminoides Ellis*

A fruit that produces chemical products such as crocins, geniposide and the phenolic compounds which are widely used in the food industry as a natural food colorants (dyes). These compounds also have high antioxidant capabilities which make this fruit valuable for medicinal uses.

# APPENDIX: MISCELLANEOUS DATA

Table A1: Values of the objective function for the GSA algorithm for various scalarization

| $\beta_1$ | $\beta_2$ | $\beta_3$ | $f_1$ | $f_2$ | $f_3$ | $X_1$ | $X_2$ | $X_3$ | Iterations |
|---|---|---|---|---|---|---|---|---|---|
| 0.1 | 0.2 | 0.7 | 5.89547 | 74.9813 | 9.84993 | 19.5 | 27.1 | 7.3491 | 46 |
| 0.1 | 0.3 | 0.6 | 6.13174 | 81.0941 | 11.9852 | 19.5 | 27.1 | 17.0239 | 183 |
| 0.1 | 0.4 | 0.5 | 5.88888 | 74.8027 | 9.78725 | 19.5 | 27.1 | 7.11013 | 22 |
| 0.1 | 0.5 | 0.4 | 6.13333 | 81.1325 | 11.9985 | 19.5 | 27.1 | 17.0981 | 183 |
| 0.1 | 0.6 | 0.3 | 6.13133 | 81.084 | 11.9817 | 19.5 | 27.1 | 17.0044 | 182 |
| 0.1 | 0.7 | 0.2 | 6.13216 | 81.1043 | 11.9887 | 19.5 | 27.1 | 17.0436 | 183 |
| 0.1 | 0.8 | 0.1 | 6.10442 | 80.4237 | 11.7523 | 19.5 | 27.1 | 15.7678 | 159 |
| 0.2 | 0.2 | 0.6 | 6.13106 | 81.0774 | 11.9794 | 19.5 | 27.1 | 16.9918 | 182 |
| 0.2 | 0.3 | 0.5 | 6.13101 | 81.0762 | 11.979 | 19.5 | 27.1 | 16.9895 | 182 |
| 0.2 | 0.4 | 0.4 | 5.88935 | 74.8154 | 9.79169 | 19.5 | 27.1 | 7.127 | 33 |
| 0.2 | 0.5 | 0.3 | 5.89061 | 74.8496 | 9.80371 | 19.5 | 27.1 | 7.17269 | 74 |
| 0.2 | 0.6 | 0.2 | 6.13329 | 81.1317 | 11.9983 | 19.5 | 27.1 | 17.0965 | 183 |
| 0.2 | 0.7 | 0.1 | 6.12652 | 80.9668 | 11.941 | 19.5 | 27.1 | 16.7801 | 179 |
| 0.3 | 0.2 | 0.5 | 5.89441 | 74.9525 | 9.83981 | 19.5 | 27.1 | 7.31041 | 52 |
| 0.3 | 0.3 | 0.4 | 5.89274 | 74.9072 | 9.82393 | 19.5 | 27.1 | 7.24972 | 56 |
| 0.3 | 0.4 | 0.3 | 5.89446 | 74.954 | 9.84033 | 19.5 | 27.1 | 7.3124 | 28 |
| 0.3 | 0.5 | 0.2 | 5.89877 | 75.0706 | 9.88126 | 19.5 | 27.1 | 7.46924 | 40 |
| 0.3 | 0.6 | 0.1 | 5.89774 | 75.0427 | 9.87148 | 19.5 | 27.1 | 7.43166 | 131 |
| 0.4 | 0.2 | 0.4 | 6.13334 | 81.1329 | 11.9987 | 19.5 | 27.1 | 17.0987 | 183 |
| 0.4 | 0.3 | 0.3 | 5.90007 | 75.1058 | 9.8936 | 19.5 | 27.1 | 7.5167 | 36 |
| 0.4 | 0.4 | 0.2 | 5.89856 | 75.065 | 9.87931 | 19.5 | 27.1 | 7.46167 | 181 |
| 0.4 | 0.5 | 0.1 | 6.13207 | 81.1019 | 11.9879 | 19.5 | 27.1 | 17.039 | 183 |
| 0.5 | 0.2 | 0.3 | 5.89744 | 75.0346 | 9.86861 | 19.5 | 27.1 | 7.42071 | 17 |
| 0.5 | 0.3 | 0.2 | 5.89236 | 74.8971 | 9.82036 | 19.5 | 27.1 | 7.23612 | 50 |
| 0.5 | 0.4 | 0.1 | 5.90698 | 75.2925 | 9.95911 | 19.5 | 27.1 | 7.76982 | 76 |
| 0.6 | 0.2 | 0.2 | 5.89499 | 74.9683 | 9.84537 | 19.5 | 27.1 | 7.33162 | 107 |
| 0.6 | 0.3 | 0.1 | 5.90615 | 75.27 | 9.95123 | 19.5 | 27.1 | 7.73926 | 31 |
| 0.7 | 0.2 | 0.1 | 5.89164 | 74.8774 | 9.81347 | 19.5 | 27.1 | 7.20983 | 73 |

Table A2: Values of the objective function for the PSO algorithm for various scalarization

| $\beta_1$ | $\beta_2$ | $\beta_3$ | $f_1$ | $f_2$ | $f_3$ | $X_1$ | $X_2$ | $X_3$ | Iterations |
|---|---|---|---|---|---|---|---|---|---|
| 0.1 | 0.3 | 0.6 | 8.52396 | 108.876 | 24.795 | 49.3379 | 72.8828 | 30.932 | 309 |
| 0.1 | 0.4 | 0.5 | 8.51943 | 108.829 | 24.7781 | 49.2946 | 72.7417 | 30.7565 | 283 |
| 0.1 | 0.5 | 0.4 | 8.52782 | 108.909 | 24.7934 | 49.484 | 72.895 | 30.7332 | 240 |
| 0.1 | 0.6 | 0.3 | 8.52153 | 108.853 | 24.782 | 49.3628 | 72.7718 | 30.7571 | 269 |
| 0.1 | 0.7 | 0.2 | 8.52156 | 108.845 | 24.7794 | 49.3387 | 72.7722 | 30.6909 | 255 |
| 0.1 | 0.8 | 0.1 | 8.5208 | 108.842 | 24.7786 | 49.3429 | 72.752 | 30.7115 | 261 |
| 0.2 | 0.2 | 0.6 | 8.52055 | 108.836 | 24.7806 | 49.2918 | 72.7752 | 30.7579 | 276 |
| 0.2 | 0.3 | 0.5 | 8.52368 | 108.87 | 24.7832 | 49.4167 | 72.7983 | 30.6892 | 245 |
| 0.2 | 0.4 | 0.4 | 8.52432 | 108.879 | 24.7888 | 49.407 | 72.8369 | 30.7838 | 266 |
| 0.2 | 0.5 | 0.3 | 8.52756 | 108.896 | 24.779 | 49.549 | 72.812 | 30.4389 | 164 |
| 0.2 | 0.6 | 0.2 | 8.52069 | 108.842 | 24.7694 | 49.4391 | 72.6776 | 30.5293 | 203 |
| 0.2 | 0.7 | 0.1 | 8.52265 | 108.862 | 24.7798 | 49.4206 | 72.7631 | 30.663 | 236 |
| 0.3 | 0.2 | 0.5 | 8.52181 | 108.85 | 24.7743 | 49.4223 | 72.7248 | 30.58 | 217 |
| 0.3 | 0.3 | 0.4 | 8.52444 | 108.878 | 24.7854 | 49.4322 | 72.8161 | 30.7055 | 244 |
| 0.3 | 0.4 | 0.3 | 8.5228 | 108.861 | 24.7803 | 49.4068 | 72.7734 | 30.6636 | 238 |
| 0.3 | 0.5 | 0.2 | 8.52059 | 108.841 | 24.7759 | 49.3695 | 72.7267 | 30.6658 | 245 |
| 0.3 | 0.6 | 0.1 | 8.52269 | 108.864 | 24.7781 | 49.4506 | 72.7456 | 30.6276 | 220 |
| 0.4 | 0.2 | 0.4 | 8.5201 | 108.834 | 24.781 | 49.2812 | 72.7724 | 30.7884 | 285 |
| 0.4 | 0.3 | 0.3 | 8.52577 | 108.888 | 24.7898 | 49.4257 | 72.8627 | 30.739 | 250 |
| 0.4 | 0.4 | 0.2 | 8.52578 | 108.895 | 24.7885 | 49.4847 | 72.8344 | 30.7215 | 233 |
| 0.4 | 0.5 | 0.1 | 8.522 | 108.856 | 24.7882 | 49.3125 | 72.8262 | 30.8662 | 301 |
| 0.5 | 0.2 | 0.3 | 8.53054 | 108.941 | 24.7943 | 49.6226 | 72.8999 | 30.6499 | 202 |
| 0.5 | 0.3 | 0.2 | 8.5202 | 108.842 | 24.7822 | 49.3185 | 72.7635 | 30.8167 | 290 |
| 0.5 | 0.4 | 0.1 | 8.52491 | 108.885 | 24.7936 | 49.386 | 72.8764 | 30.8617 | 286 |
| 0.6 | 0.2 | 0.2 | 8.5251 | 108.888 | 24.7777 | 49.565 | 72.7498 | 30.5274 | 183 |
| 0.6 | 0.3 | 0.1 | 8.52539 | 108.883 | 24.7794 | 49.5095 | 72.7831 | 30.5408 | 189 |
| 0.7 | 0.2 | 0.1 | 8.52065 | 108.84 | 24.7806 | 49.3153 | 72.7677 | 30.7582 | 275 |

Table A3: Values of the objective function for the HPSO algorithm for various scalarization

| $\beta_1$ | $\beta_2$ | $\beta_3$ | $f_1$ | $f_2$ | $f_3$ | $X_1$ | $X_2$ | $X_3$ | Iterations |
|---|---|---|---|---|---|---|---|---|---|
| 0.1 | 0.2 | 0.7 | 8.54683 | 109.903 | 24.6649 | 57.699 | 72.5172 | 36.9105 | 76 |
| 0.1 | 0.3 | 0.6 | 8.56443 | 109.547 | 24.6564 | 55.4598 | 72.616 | 29.3023 | 55 |
| 0.1 | 0.4 | 0.5 | 8.56161 | 109.771 | 24.8042 | 54.6208 | 72.7013 | 32.4585 | 67 |
| 0.1 | 0.5 | 0.4 | 8.54641 | 109.35 | 24.7936 | 52.1086 | 72.5603 | 30.8118 | 84 |
| 0.1 | 0.6 | 0.3 | 8.56848 | 109.741 | 24.6144 | 57.4313 | 72.8477 | 30.215 | 88 |
| 0.1 | 0.7 | 0.2 | 8.5439 | 109.74 | 24.5534 | 58.3004 | 72.891 | 40.5242 | 71 |
| 0.1 | 0.8 | 0.1 | 8.55699 | 109.725 | 24.59 | 57.8938 | 72.6076 | 30.9019 | 59 |
| 0.2 | 0.2 | 0.6 | 8.55445 | 109.542 | 24.7869 | 53.4146 | 72.5551 | 31.1148 | 57 |
| 0.2 | 0.3 | 0.5 | 8.57352 | 109.556 | 24.6483 | 55.5045 | 72.843 | 28.7859 | 42 |
| 0.2 | 0.4 | 0.4 | 8.54243 | 109.881 | 24.6273 | 58.1635 | 72.3826 | 36.6553 | 77 |
| 0.2 | 0.5 | 0.3 | 8.55299 | 109.679 | 24.8298 | 53.7879 | 72.6255 | 32.9866 | 71 |
| 0.2 | 0.6 | 0.2 | 8.54623 | 109.923 | 24.6032 | 58.693 | 72.5267 | 36.6219 | 58 |
| 0.2 | 0.7 | 0.1 | 8.56719 | 109.715 | 24.8006 | 54.1888 | 72.8216 | 31.4303 | 56 |
| 0.3 | 0.2 | 0.5 | 8.55619 | 109.907 | 24.7863 | 55.9271 | 72.7838 | 36.1053 | 74 |
| 0.3 | 0.3 | 0.4 | 8.54763 | 109.783 | 24.7766 | 55.363 | 72.379 | 34.124 | 70 |
| 0.3 | 0.4 | 0.3 | 8.57379 | 109.544 | 24.6937 | 54.6109 | 72.8529 | 28.9895 | 63 |
| 0.3 | 0.5 | 0.2 | 8.5531 | 109.028 | 24.3727 | 57.2096 | 72.6254 | 26.2125 | 60 |
| 0.3 | 0.6 | 0.1 | 8.54398 | 108.982 | 24.748 | 50.1817 | 72.8661 | 29.2429 | 68 |
| 0.4 | 0.2 | 0.4 | 8.55019 | 109.536 | 24.7396 | 54.2021 | 72.3286 | 30.8219 | 70 |
| 0.4 | 0.3 | 0.3 | 8.54072 | 109.474 | 24.8598 | 52.2739 | 72.6808 | 33.3906 | 58 |
| 0.4 | 0.4 | 0.2 | 8.56125 | 108.326 | 24.1797 | 55.3454 | 72.81 | 22.7159 | 59 |
| 0.4 | 0.5 | 0.1 | 8.55578 | 109.406 | 24.7608 | 52.8167 | 72.6043 | 30.0057 | 51 |
| 0.5 | 0.2 | 0.3 | 8.55857 | 109.167 | 24.4196 | 57.2167 | 72.7171 | 26.7412 | 71 |
| 0.5 | 0.3 | 0.2 | 8.54968 | 109.951 | 24.5644 | 59.4341 | 72.6607 | 35.4821 | 59 |
| 0.5 | 0.4 | 0.1 | 8.5675 | 109.929 | 24.8 | 55.6416 | 72.8689 | 33.5713 | 84 |
| 0.6 | 0.2 | 0.2 | 8.5503 | 109.192 | 24.7651 | 51.3036 | 72.7476 | 29.6299 | 55 |
| 0.6 | 0.3 | 0.1 | 8.55 | 109.56 | 24.7846 | 53.6643 | 72.4305 | 31.5473 | 56 |
| 0.7 | 0.2 | 0.1 | 8.5607 | 108.642 | 24.3953 | 53.428 | 72.5883 | 24.4832 | 36 |